# Synchronized Detection and Recovery of Steganographic Messages with Adversarial Learning

Haichao Shi[1,2] and Xiao-Yu Zhang[1]

[1] Institute of Information Engineering, Chinese Academy of Sciences, Beijing, China
[2] School of Cyber Security, University of Chinese Academy of Sciences, Beijing, China
`zhangxiaoyu@iie.ac.cn`

**Abstract.** In this work, we mainly study the mechanism of learning the steganographic algorithm as well as combining the learning process with adversarial learning to learn a good steganographic algorithm. To handle the problem of embedding secret messages into the specific medium, we design a novel adversarial modules to learn the steganographic algorithm, and simultaneously train three modules called generator, discriminator and steganalyzer. Different from existing methods, the three modules are formalized as a game to communicate with each other. In the game, the generator and discriminator attempt to communicate with each other using secret messages hidden in an image. While the steganalyzer attempts to analyze whether there is a transmission of confidential information. We show that through unsupervised adversarial training, the adversarial model can produce robust steganographic solutions, which act like an encryption. Furthermore, we propose to utilize supervised adversarial training method to train a robust steganalyzer, which is utilized to discriminate whether an image contains secret information. Numerous experiments are conducted on publicly available dataset to demonstrate the effectiveness of the proposed method.

## 1 Introduction

Steganography aims to conceal a payload into a cover object without affecting the sharpness of the cover object. The image steganography is the art and science of concealing covert information within images, and is usually achieved by modifying image elements, such as pixels or DCT coefficients. Steganographic algorithms are designed to hide the secret information within a cover message such that the cover message appears unaltered to an external adversary. On the other side, steganalysis aims to reveal the presence of secret information by detecting the abnormal artefacts left by data embedding and recover the secret information of the carrier object. For a long period of time, many researchers have been involved in developing new steganographic systems. Meanwhile, the development of steganalytic tools are also started growing. We have found that the relationship between these two aspects is similar with adversarial learning, so we leverage the ideas from adversarial learning to optimize them simultaneously.

Adversarial learning is based on the game theory, and is combined with unsupervised way to jointly train the model. In our previous work [19], we proposed a novel strategy of secure steganograpy based on generative adversarial networks to generate suitable and secure covers for steganography, in which the decoding process was not considered. In this paper, we not only encode the secret messages into the images, but also decode it utilizing the network. Then, we utilize the steganalysis network to detect the presence of hidden messages. Through unsupervised training, the generator plays the role of a sender, which is utilized to generate steganographic images as real as possible. As the discriminator plays the role of a receiver, it not only differentiates real images and generated images, but also extracts the secret messages. And the steganalysis network, as a listener of the whole process, incorporates supervised learning with adversarial training to compete against state-of-the-art steganalysis methods.

In summary, this paper makes the following contributions:

(1) We incorporate the generative adversarial network into steganography, which proves to be a good way of encryption.

(2) We integrate the unsupervised learning method and supervised learning method to train the generator and steganalyzer respectively, and receive robust steganographic techniques in unsupervised manner.

(3) We also utilize the discriminative network to extract the secret information. Experiments are conducted on widely used datasets, to demonstrate the advantages of the proposed method.

The rest of the paper is structured as follows. In Section 2, we discuss the related work of steganography, adversarial networks. In Section 3, we elaborate the proposed method. In Section 4, experiments are conducted to demonstrate the effectiveness of the proposed method. In Section 5, we draw conclusions.

## 2 Related work

### 2.1 Steganography

The image-based steganography algorithm can be split into two categories. The one is based on the spatial domain, the other is based on the DCT domain. In our work, we mainly focus on the spatial domain steganography. The state-of-the-art steganographic schemes concentrate on embedding secret information within a medium while minimizing the perturbations within that medium. On the contrary, steganalysis is to figure out whether there is secret information or not in the medium.

Least Significant Bit (LSB) [12] is one of the most popular embedding methods in spatial domain steganography. If LSB is adopted as the steganography method, the statistical features of the image are destroyed. And it is easy to detect by the steganalyzer. For convenience and simple implementation, the LSB algorithm hides the secret to the least significant bits in the given image's channel of each pixel. Mostly, the modification of the LSB algorithm is called ±1-embedding. It randomly adds or subtracts 1 from the channel pixel, so the last bits would match the ones needed.

Besides the LSB algorithm, some sophisticated steganographic schemes choose to use a distortion function which is used for selecting the embedding localization of the image. This type of steganography is called the content-adaptive steganography. The minimization of the distortion function between the cover image $C$ and the steganographic image $S$ is usually required. These algorithms are the most popular and the most secure image steganography in spatial domain, such as HUGO (Highly Undetectable steGO) [4], WOW (Wavelet Obtained Weights) [2], S-UNIWARD (spatial universal wavelet relative distortion) [3], etc.

$$d(C,S) = f(C,S) * |C - S| \qquad (1)$$

where $f(C,S)$ is the cost of modifying a pixel, which is variable in different steganographic algorithms.

HUGO is a steganographic scheme that defines a distortion function domain by assigning costs to pixels based on the effect of embedding some information within a pixel. It uses a weighted norm function to represent the feature space. WOW is another content-adaptive steganographic method that embeds information into a cover image according to textural complexity of regions. It is shown in WOW that the more complex the image region is, the more pixel values will be modified in this region. S-UNIWARD introduces a universal distortion function that is independent of the embedded domain. Despite the diverse implementation details, the ultimate goals are identical, i.e. they are all devoted to minimize this distortion function, to embed the information into the noise area or complex texture, and to avoid the smooth image coverage area.

### 2.2 Adversarial Learning

In recent years, Generative Adversarial Networks (GANs) have been successfully applied to image generation tasks. The method that generative adversarial networks generate images can be classified in two categories in general. The first is mainly exploring image synthesis tasks in an unconditioned manner that generates synthetic images without any supervised learning schemes. Goodfellow et al. [7] propose a theoretical framework of GANs and utilize GANs to generate images without any supervised information. However, the early GANs has somewhat noisy and blurry results and sometimes the gradient will be vanished when training the networks. Later, Radford et al. [15] propose a deep convolutional generative adversarial networks (DCGANs) for unsupervised representation. To solve the situation of gradient vanishing, WGAN [14] is proposed using the Wasserstein distance instead of the Jensen-Shannon divergence, to make the data set distribution compared with the learning distribution from G.

Another direction of image synthesis with GANs is to synthesize images by conditioning on supervised information, such as text or class labels. The Conditional GAN [16] is one of the works that develop a conditional version of GANs by additionally feeding class labels into both generator and discriminator of GANs. Info-GAN [17] introduces a new concept, which divides the input noise z into two parts, one is the continuous noise signal that cannot be explained, and the other is called C. Where C represents a potential attribute that can be interpreted as a facial expression, such as the color of eyes, whether with glasses or not, etc. in the facial tasks. Recently, Reed et al.

[18] utilize GANs for image synthesis using given text descriptions, enabling translation from character level to pixel level.

Adversarial learning has been applied to steganographic and cryptographic problems. In Abadi's [1] work, they integrate two neural networks to adversarial game to encrypt a secret message. In which game the discriminator can be deceived. In this paper, we are devoted to training a model that can learn a steganographic technique by itself making full use of the discriminator and recovering the secret message synchronously.

## 3   Adversarial Steganography

This section mainly discusses our proposed adversarial steganographic scheme. In section 3.1, we elaborate the network architecture of our model. Then, we formulate the problem of steganography and steganalysis to point out the foundation of the proposed steganographic scheme.

### 3.1   Network Architecture

Our model consists of three components. The generator is used to generate steganographic images. Besides discriminating between real images and generated images, the discriminator is utilized to extract the secret messages. And the steganalyzer is utilized to distinguish covers from steganographic images.

**Generator.** The generator receives the cover images and secret messages. Before learning the distribution of the images, the secret messages are first embedded into the cover image. The generated steganographic images are then put into the generator. As shown in Fig.1, we use a fully connected layer, which can cover anywhere of the image not merely a fixed region. Then four fractionally-strided convolution layers, and finally a Hyperbolic tangent function layer. The architecture is as shown in **Fig. 1.**

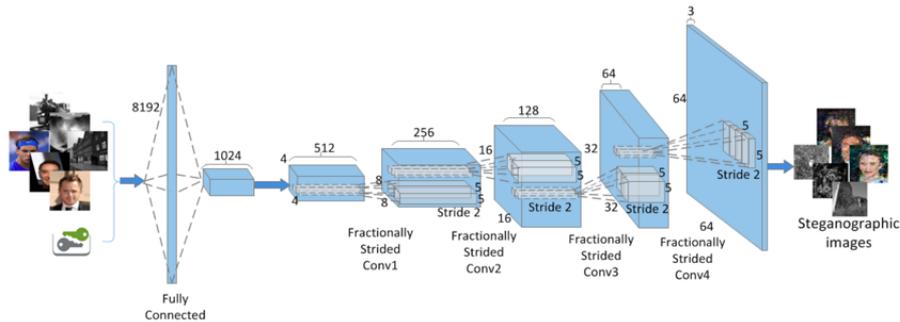

**Fig. 1.** The architecture of the generator

**Discriminator.** The discriminator is mainly responsible for extracting the secret messages. Besides, it can also help to optimize the visual quality of the generated im-ages. We use four convolutional layers, and a fully connected layer. In detail, we additionally add a decoding function at the end of the network to extract the secret messages. The decoding function is acted as an interface between the steganographic images and the secret messages. It analyzes the modification of pixels in images, and recovers the message. The architecture is as shown in **Fig. 2**.

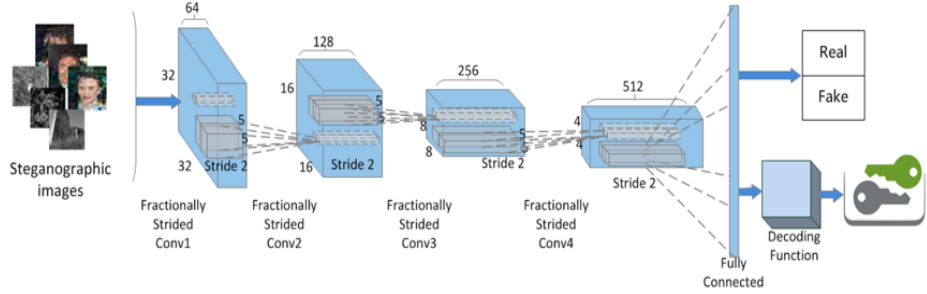

**Fig. 2.** The architecture of the discriminator

**Steganalyzer.** The steganalyzer's inputs are both covers and steganographic images. The steganalyzer's architecture is similar to the discriminator, as is illustrated in **Fig. 3**. With distinct, we first use a predefined high-pass filter to make a filtering operation, which is mainly for steganalysis. And then we use four convolutional layers. Finally, we use a classification layer, which includes several fully connected layers.

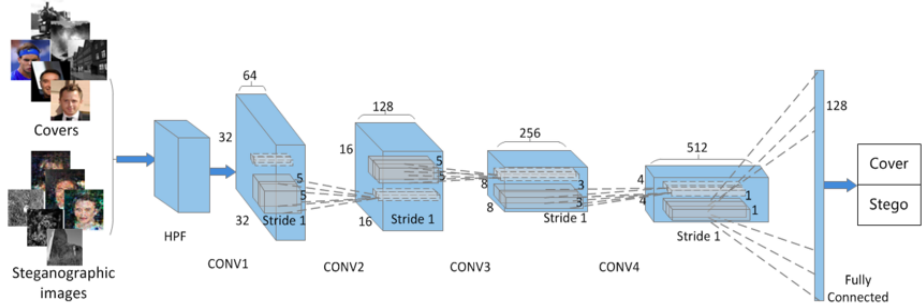

**Fig. 3.** The architecture of the steganalyzer

### 3.2 Optimization Objective

Our training scheme includes three parties: generator, discriminator and steganalyzer. In this game, the generator conveys a secret medium to the discriminator, which the steganalyzer attempts to eavesdrop the conveying process and find out whether there are some special information containing in the process. Generally speaking, the generator generates a steganographic image within the cover image embedded with a secret message. Then generator passes it to the next network which we call discriminator, concentrating on decoding and recovering the message. When an image is input to the steganalyzer, a confidence probability will be output by steganalyzer on how much likely the image is hidden with a secret message.

For the whole process of information conveying is transparent to the discriminator, so the decoding can be easily achieved. In the game of the three parties, the generator is trained to learn to produce a steganographic image such that the discriminator can decode and recover the secret message, and such that the steganalyzer can output a better probability whether it is a cover or a steganographic image.

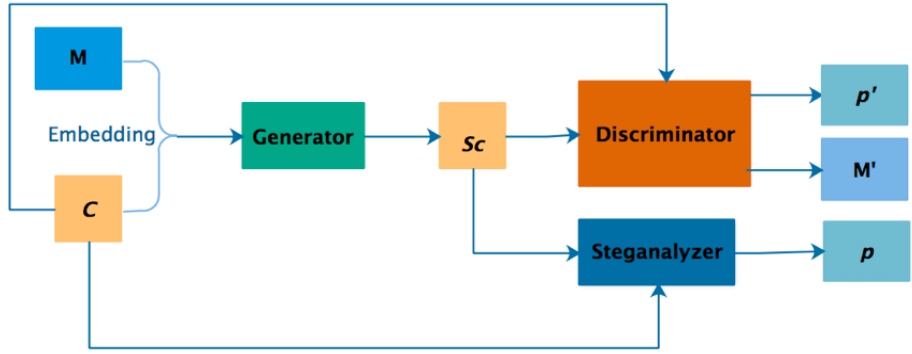

**Fig. 4.** The whole architecture of the training game. $M$ and $M'$ represents the secret message and extracted message from the steganographic image respectively, $C$ represents the cover image, and $S_C$ represents the steganographic image. $p$ is the probability of classifying the covers and steganographic images. $p'$ is the probability of classifying the real images and generated images.

The whole architecture can be depicted in **Fig. 4**, the generator receives a cover image, $C$, and a secret message, $M$. Combined with the above two inputs, it outputs a steganographic image, $S_C$, which is simultaneously given to the discriminator and steganalyzer. Once receiving the steganographic image, the discriminator decodes it and attempts to recover the message. For the game is under generative adversarial networks, the discriminator can also improve the visual quality of the generated images. In addition to the steganographic images, the cover images are also input into the steganalyzer, whose output is a probability to classify the steganographic images and covers.

We let $\omega_G$, $\omega_D$, $\omega_S$ denote the parameters of generator, discriminator and steganalyzer respectively. We utilize $G(\omega_G, M, C)$ represents generator's output given an image $C$ and secret message $M$, $D(\omega_D, S_C)$ for discriminator's output on steganographic images $S_C$, and $S(\omega_S, C, S_C)$ for steganalyzer's output on covers and steganographic images. Then we use $L_G$, $L_D$, $L_S$ represent the loss of three networks respectively. And we can calculate as follows:

$$D(\omega_D, S_C) = D(\omega_D, G(\omega_G, M, C)) \qquad (2)$$

$$S(\omega_S, C, S_C) = S(\omega_S, C, G(\omega_G, M, C)) \qquad (3)$$

where the $D(\omega_D, S_C)$ is output of the discriminator, $S(\omega_S, C, S_C)$ is the output of the steganalyzer.

For the generator's loss, we correspond it with the other two networks. And we add constraints to normalize the loss.

$$L_G(\omega_G, M, C) = \lambda_G \cdot d(C, S_C) + \lambda_D \cdot L_D(\omega_D, S_C) + \lambda_S \cdot L_S(\omega_S, C, S_C) \quad (4)$$

where $d(C, S_C)$ is the Euclidean distance between the covers and the steganographic images. And $\lambda_G, \lambda_D, \lambda_S$ are the calculation loss weights of the generator, discriminator and steganalyzer respectively. These variables are varying from [0,1].

Also, we set discriminator's loss to be the norm form. Here, we use the $\ell_2$ norm.

$$L_D(\omega_G, \omega_D, M, C) = ||M, D(\omega_D, S_C)||_2 = ||M, D(\omega_D, G(\omega_G, M, C))||_2 = ||M, M'||_2 \quad (5)$$

where the form $||M, M'||_2$ is the distance between $M$ and $M'$.

And we set steganalyzer's loss to be binary cross-entropy of logistic regression.

$$L_S(\omega_G, \omega_S, C, S_C) = -\frac{1}{n}\sum_x y \ln S(\omega_S, C, S_C) + (1-y)\ln(1 - S(\omega_S, C, S_C)) \quad (6)$$

where $y = 1$ if $x = C$, which means classifying the images as covers, $y = 0$ if $x = S_C$, which means the steganographic images are identified.

We apply adversarial learning techniques to discriminative tasks to learn the steganographic algorithm. Under unsupervised adversarial training, the model can produce steganographic techniques, which act like an encryption.

## 4    Experiments

Experiments are conducted on two image datasets: celebrity faces in the wild (CelebA) [8] consisting of 202,599 images and BossBase, which is a dataset of 10000 grayscale images of various scenes. Our datasets are formed of 64 × 64 pixel images. For the choice of $M$, we concatenate a random paragraph message to generalize our model to random message, with each sample of each dataset. And we consider different kinds of embedding rates, which vary from 0.1bpp (bits per pixel) to 0.4bpp. Generally, most steganographic algorithms can successfully hide secret message approximately at 0.4bpp. All experiments were performed in TensorFlow [9], on a workstation with a Titan X GPU.

### 4.1    CelebA dataset

CelebA dataset contains 202,599 images of celebrity faces from 10,177 unique identities. We train the model using RMSProp optimization algorithm with the learning rate of $2 \times 10^{-4}$. At each batch we train either the generator, discriminator, or steganalyzer alternatively.

From the results, we can see that there is no noticeable image quality decrease between 0.1bpp and 0.4bpp. As is shown in **Fig. 5**. And we also calculate the PSNR (Peak Signal to Noise Ratio) to evaluate the steganographic effect. As is shown in **Fig. 6**.

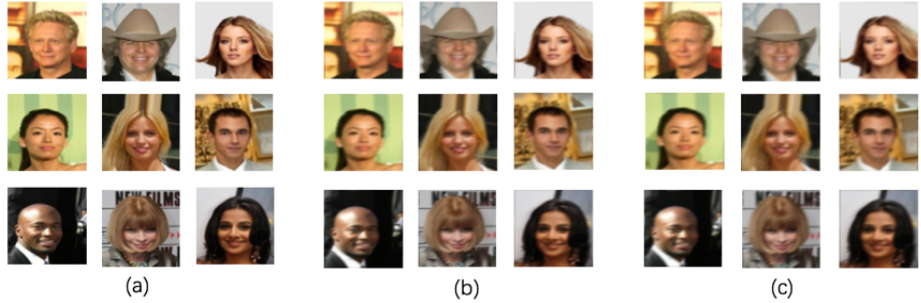

**Fig. 5.** The above are the covers and steganographic images of CelebA dataset. (a) represents the covers, (b) represents the steganographic images with the embedding rates of 0.1bpp, (c) represents the steganographic images with the embedding rates of 0.4bpp.

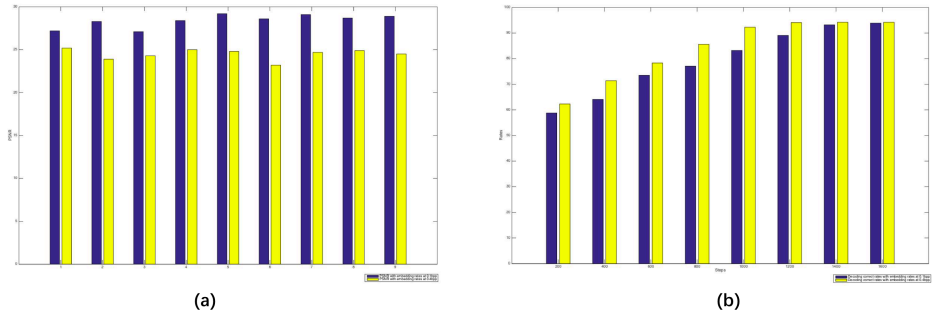

**Fig. 6.** (a) represents the contrast of PSNR of embedding rates at 0.1bpp and 0.4bpp. (b) represents the contrast of successful decoding rates between 0.1bpp and 0.4bpp. The blue bar represents the results at 0.1bpp, while the yellow bar represents the results at 0.4bpp.

From the charts, we can see that the discriminator can hardly decode the secret messages in the first few rounds of training. For the visual quality of the generated images is low and the discriminator is randomly guessing the message essentially. After 800 training steps, the discriminator can decode the message correctly with an average success of 90% at 0.4bpp. After 1,000 training steps, the discriminator approaches to convergence gradually and decodes the message with success rate above 95%. We can also conclude that the discriminator performs better under embedding rate at 0.4bpp than 0.1bpp when decoding the messages.

### 4.2 BossBase dataset

In addition to the experiments on the CelebA dataset, we also trained our model on the BossBase dataset, which is a standard steganography dataset. For the images from this dataset do not come from a single distribution, so the dataset can perform worse than the experiments on the CelebA dataset.

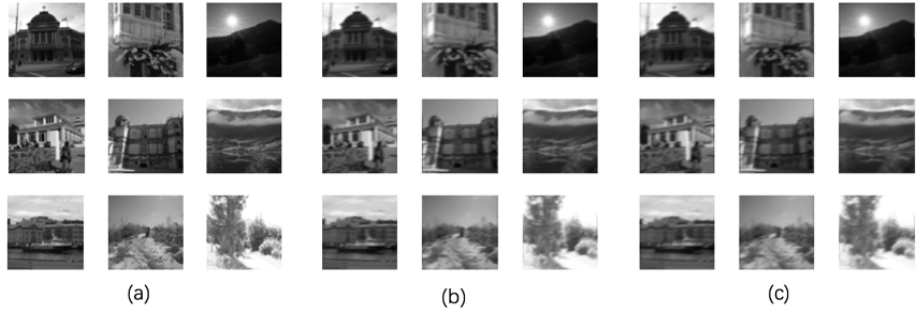

**Fig. 7.** The above are the covers and steganographic images of BossBase dataset. (a) represents the covers, (b) represents the steganographic images with the embedding rates of 0.1bpp, (c) represents the steganographic images with the embedding rates of 0.4bpp.

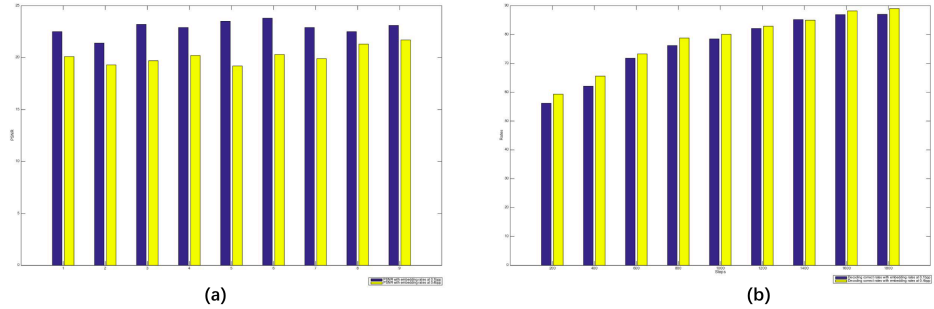

**Fig. 8.** (a) represents the contrast of PSNR of embedding rates at 0.1bpp and 0.4bpp. (b) represents the contrast of successful decoding rates between 0.1bpp and 0.4bpp. The blue bar represents the results at 0.1bpp, while the yellow bar represents the results at 0.4bpp.

It is shown in **Fig. 8** that the contrast of PSNR and successful decoding rates at 0.1bpp and 0.4bpp. While the decoding rates suffers slightly from the image quality, the discriminator is still able to learn to decode the message successfully. This also implies the effectiveness of the generator, which is able to learn a good steganographic algorithm.

We can conclude from the two groups of experiments that for the datasets with distinct distribution, different embedding rates can influence the similarity between the original images and the embedded images.

### 4.3 Comparison with the state-of-the-arts

We train in such a way that the steganalyzer can detect the covers and steganographic images with low rates. Thus, on the basis of this experiment, we additionally conduct another experiment compared with the state-of-the-art steganographic algorithms. Which also shows competitive results.

For both datasets, we compare our scheme against steganographic algorithms HUGO and S-UNIWARD. As is shown in **Table 1**, our steganalyzer performs competitively. By training steganalyzer in a supervised fashion, the experiments show that it has the capacity to become a strong steganalyzer, competing against established techniques like GNCNN, and so is a good choice for the steganalyzer. Furthermore, for both the CelebA and BossBase datasets, we use 10,000 samples and split them in half, creating a training set and a test set. And we then trained the generator on the training set. Following this, we then created an additional 5,000 steganographic images utilizing each steganography algorithm. Thus, we can make the cover set and steganographic images set consist of 5,000 images respectively.

**Table 1.** Accuracy of distinguishing between cover and steganographic images for the GNCNN and Steganalyzer at an embedding rate of 0.4bpp.

| Steganographic Algorithm | CelebA Steganalyzer | | BossBase Steganalyzer | |
|---|---|---|---|---|
| | GNCNN | Steganalyzer | GNCNN | Steganalyzer |
| HUGO | 0.93 | 0.91 | 0.72 | 0.68 |
| S-UNIWARD | 0.91 | 0.89 | 0.75 | 0.71 |
| Generator | 0.94 | 0.92 | 0.83 | 0.82 |

As demonstrated in the experiments, the steganalyzer performs competitively against the GNCNN [10], and the generator also performs well against other steganographic techniques. The experimental data shows that on the one hand, the generated images are good at visual quality. On the other hand, the generated images are harder to detect, which shows the security of the generated images as covers.

## 5 Conclusion

In this paper, an adversarial steganography architecture with generative adversarial networks is proposed. On the basis of generative adversarial networks, we leverage the adversarial structure to form an effective steganographic method. Encouraging results are received from experiments conducted on the widely used datasets in comparison with several state-of-the-art methods. It is also worth to study the adoption of the adversarial steganography architecture in adaptive steganographic algorithm. Furthermore, it is interesting to resort the networks to a game-theoretic formulation when we characterize the interplay between the steganographer and the steganalyst.